\documentclass{article}
\usepackage{ijcai26}

\usepackage{times}
\usepackage{soul}
\PassOptionsToPackage{hyphens}{url}
\usepackage{url}
\usepackage[hidelinks]{hyperref}
\usepackage[utf8]{inputenc}
\usepackage[small]{caption}
\usepackage{graphicx}
\usepackage{amsmath}
\usepackage{amsthm}
\usepackage{booktabs}
\usepackage{algorithm}
\usepackage{algorithmic}
\usepackage[switch]{lineno}

\usepackage{caption}
\usepackage{newfloat}
\usepackage{listings}
\usepackage{xcolor}
\usepackage{algorithm}
\usepackage{algorithmic}
\usepackage{newfloat}
\usepackage{amsfonts}
\usepackage{listings}
\usepackage{booktabs}
\usepackage{multirow}
\usepackage{amsmath}
\usepackage[table]{xcolor}

\title{RLCascadeRouter: Quality-Estimator-Free Cascade Routing via Reinforcement Learning}

\author{
Shihong Huang$^{1}$,
Shengjie Wang$^{1}$,
Hong Ma$^{1}$,
Zhou Xu$^{2,}$\thanks{
Corresponding authors.}\\
\affiliations
$^1$Polytechnic Institute, Zhejiang University, Hangzhou, China\\
$^2$Department of Logistics and Maritime Studies, Faculty of Business, The Hong Kong Polytechnic University, Hong Kong, China
}

\begin{document}

\maketitle

\begin{abstract}
The growing ecosystem of large language models (LLMs) offers huge potential to optimize performance-cost trade-offs. However, their heterogeneous capabilities and inference costs make efficiently routing queries a significant challenge.
Existing paradigms are inflexible: one-shot routers commit before observing responses, whereas conventional cascades stop adaptively but follow a fixed model order. Cascade routing removes both restrictions by reconsidering whether to stop or invoke another model after each response. Current methods use a predict-then-optimize pipeline estimating response quality and future model utility. However, prediction loss for quality or utility is not equivalent to routing-decision loss. A lower prediction error does not necessarily yield a better action; a small boundary-crossing error can reverse a ``stop'' or model-selection decision. Therefore, we propose RLCascadeRouter, a quality-estimator-free framework that formulates cascade routing as a Markov decision process with actions comprising ``stop'' and model selection. It uses trajectory returns and advantages to directly optimize the performance-cost objective. Its Cascade Policy Network models candidate complementarity for model selection and remaining-action value for stopping, eliminating independent post-hoc response-quality estimators. Evaluated across ten LLMRouterBench benchmarks with thirteen LLMs, RLCascadeRouter outperforms strong baselines and achieves superior performance-cost trade-offs. It incorporates unseen models without retraining, and ablation studies validate both policy components.
\end{abstract}


\section{Introduction}
The rapid development of large language models (LLMs) has created a diverse ecosystem of models with heterogeneous capabilities and inference costs. While large, general-purpose models often provide strong average performance, invoking them for every query can be prohibitively expensive. Conversely, smaller or specialized models may solve many queries at a substantially lower cost, but their performance can vary considerably across tasks and instances. This heterogeneity motivates LLM model selection: given a query and a pool of candidate models, the system should determine which model or sequence of models can produce a satisfactory answer while controlling the inference cost.~\cite{varangot2026doing,moslem2026dynamic}

\begin{figure*}[!ht]
    \centering
    \includegraphics[width=\linewidth]{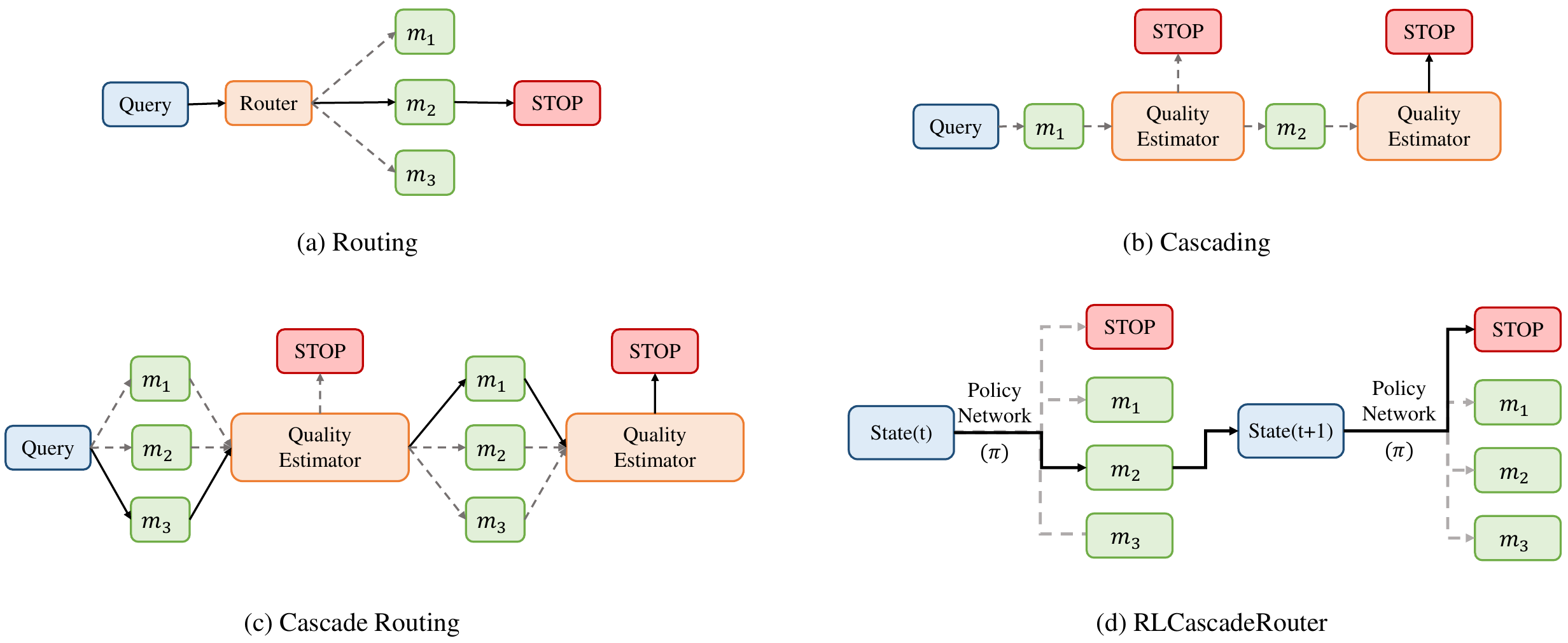}
    \caption{Comparison of LLM model-selection structures: (a) one-shot routing, (b) fixed-order cascading with quality-based stopping, (c) estimator-based cascade routing, and (d) RLCascadeRouter, which jointly learns ``stop'' and model actions through a unified MDP policy.}
    \label{fig:framework}
\end{figure*}

Existing model-selection methods mainly follow two paradigms: \emph{routing} and \emph{cascading}~\cite{moslem2026dynamic}. As illustrated in Figure~\ref{fig:framework}(a), routing makes a one-shot decision that assigns each query to a candidate model, typically based on query features and the estimated capabilities and costs of the candidates~\cite{zhuang2025embedllm,feng2025graphrouter}. Cascading, shown in Figure~\ref{fig:framework}(b), arranges models in a predefined sequence and invokes them successively until the current response is considered satisfactory~\cite{chen2024frugalgpt,aggarwal2024automix,gupta2024language}. Despite their utility, both paradigms restrict the model-selection process. Routing can adapt the selected model but cannot revise its decision after observing the generated response. Cascading can adapt the number of model calls, but its fixed order prevents query-specific skipping or reordering. A less restrictive strategy that supports both dynamic model selection and dynamic stopping could therefore provide a stronger performance-cost trade-off.

Dekoninck et al.~\shortcite{dekoninck2025unified} take an important step toward this goal with \emph{cascade routing}, illustrated in Figure~\ref{fig:framework}(c). Their framework dynamically selects between stopping and future model combinations, generalizing both one-shot routing and fixed-order cascading. However, this framework remains critically dependent on quality estimation: ex-ante estimates assess the value of models, while post-hoc estimates determine whether the current response is sufficient to stop. This predict-then-optimize design creates a Prediction-Decision Mismatch because prediction loss and downstream decision loss are not equivalent. A small estimation error that crosses an action boundary can reverse a stopping or model-selecting decision, whereas a larger error that preserves the action ordering may have no decision consequence. Consequently, improving the quality estimator does not necessarily improve the resulting routing policy.

Therefore, we propose \textbf{RLCascadeRouter}, a quality-estimator-free cascade-routing framework based on reinforcement learning. As shown in Figure~\ref{fig:framework}(d), we formulate cascade routing as a Markov decision process (MDP) with an action space that jointly contains ``stop'' and all previously unselected models. Rather than predicting response qualities and subsequently converting them into routing decisions, RLCascadeRouter learns action preferences directly from trajectory-level performance-cost feedback. Its Cascade Policy Network captures two relationships required by cascade routing: capability complementarity among the remaining models and the relative value of the current state versus the remaining actions. Consequently, ``stop'' and model selection are learned jointly under the final objective, without invoking an independently trained quality estimator or calibrated stopping threshold.

Our main contributions are threefold:
\begin{itemize}
    \item We demonstrate the Prediction--Decision Mismatch, and formulate cascade routing as an MDP that directly optimizes the final performance-cost objective through trajectory returns and advantages.
    \item We develop a Cascade Policy Network for quality-estimator-free cascade routing. Its Complementarity Encoder models the relative values of the remaining models, while its Value-Aware Stopper compares the current state with the action context to jointly learn when to stop and which model to invoke next.
    \item We evaluate RLCascadeRouter on LLMRouterBench~\cite{li-etal-2026-llmrouterbench}. The results demonstrate strong overall performance, favorable performance-cost trade-offs, generalization to unseen models without policy retraining, and the effectiveness of the proposed policy components.
\end{itemize}

\section{Related Work}
\subsection{LLM Routing and Cascading}
LLM routing seeks to exploit differences in model capability and inference cost by assigning each query to a suitable model~\cite{huang2025routereval,hu2024routerbench}. One of the most common applications of routing is model selection for natural language input queries with known answers~\cite{chuang2024learning,liu2024optllm,nguyen2024metallm,jang2023exploring}. Early approaches primarily execute this decision prior to generation. HybridLLM~\cite{ding2024hybridllm} predicts query difficulty and routes between a weaker and a stronger model according to a configurable quality target. RouteLLM~\cite{ong2025routellm} learns a strong-versus-weak routing boundary from human preference data and improves transfer through data augmentation. Subsequent methods extend routing to larger and dynamic model pools. RouterDC~\cite{chen2024routerdc} learns query and model representations with dual contrastive objectives, while GraphRouter~\cite{feng2025graphrouter} represents tasks, queries, and LLMs as a heterogeneous graph and predicts performance-cost attributes for query--model edges. Avengers-Pro~\cite{zhang2025avengerspro} instead uses query clustering and cluster-level model profiles to select a model under different performance-cost preferences. Despite their different representations and learning objectives, these methods ultimately commit to one model before observing its response.

Cascading introduces post-generation feedback by invoking models successively and stopping when the current answer is considered reliable. FrugalGPT~\cite{chen2024frugalgpt} learns both an ordered LLM cascade and response scorers that determine whether to return an answer or continue. Mixture-of-Thoughts~\cite{yue2024mixturethoughts} uses repeated reasoning traces and response consistency to decide whether a query should be escalated to a stronger model. Select-then-Route~\cite{shah2025selectthenroute} first constructs a task-relevant candidate pool and then executes a confidence-based cascade from cheaper to more capable models. Such methods can vary the number of calls, but usually retain a predetermined escalation order and delegate the stopping decision to a confidence score, consistency rule, or quality estimator.

Routing and cascading therefore provide complementary but incomplete forms of adaptivity. Single-step routers can choose among a broad model pool but cannot revise the decision after generation, whereas conventional cascades can reconsider an answer but cannot freely choose the next model. In contrast, RLCascadeRouter treats ``stop'' and all unselected candidate models as peer actions at every decision step, enabling both the cascade order and its depth to depend on the query and observed responses.

\subsection{Reinforcement-Learning-Based Router}
Several works formulate model selection as adaptive learning or sequential control~\cite{shao2025route,zheng2024mixllm,zhang2025routerr1,panda2025adaptive}. AutoMix~\cite{aggarwal2024automix} obtains an answer from a smaller model, estimates its reliability through few-shot self-verification, and uses a POMDP-based router to accept the answer or escalate within an ordered model hierarchy. PILOT~\cite{panda2025adaptive} formulates routing as a contextual bandit and updates query--model affinities from online feedback under a budget policy; however, each query still produces a single model-selection action. TREACLE~\cite{zhang2024treacle} formulates model-and-prompt selection as a constrained MDP and trains a DQN policy using query embeddings, response consistency, re-query counts, and remaining budget. Its actions return the current answer, repeat the current model--prompt pair, or advance to the next pair in a global cascade ordered by accuracy--cost ratio. Dekoninck et al.~\shortcite{dekoninck2025unified} provide a more general theoretical treatment. Their cascade-routing formulation dynamically compares stopping with future supermodels, establishes optimal strategies for routing and cascading, and identifies ex-ante and post-hoc quality estimation as critical to practical performance. Its implementation nevertheless depends on estimating the utility of unqueried models and generated responses and on evaluating a combinatorial family of future continuations. Router-R1~\cite{zhang2025routerr1} takes a different RL-based approach: it instantiates the router as a capable LLM that interleaves internal \emph{think} actions with multi-round \emph{route} actions and aggregates the resulting responses into a final answer. This design targets agentic reasoning and multi-model aggregation rather than lightweight cascade control.

\section{RLCascadeRouter}
We present our methodology in three parts.
Section~\ref{sec:decision-misalignment} formalizes the mismatch between quality
prediction and routing decisions. Section~\ref{sec:mdp} defines the MDP formulation. Finally, Section~\ref{sec:hierarchical-policy} introduces the Cascade Policy Network used to approximate the resulting stopping and selection policy.

\subsection{Prediction--Decision Mismatch}
\label{sec:decision-misalignment}

\paragraph{Prediction versus decision.}
Estimator-based cascade routing follows a predict-then-optimize pipeline. At a state $s$, it first predicts the performance-cost utilities of the current and future options and then selects the option with the largest predicted utility.
Let $u(s)\in\mathbb R^{|\mathcal A(s)|}$ denote the true long-term utility
vector of the valid actions, and let $\widehat u(s)$ be its prediction. The true
and predicted decisions are
\begin{equation}
    a^*(s)\in\arg\max_a u_a(s),
    \qquad
    \widehat a(s)\in\arg\max_a\widehat u_a(s).
\end{equation}
A prediction objective may minimize
$L_{\mathrm{pred}}=\|\widehat u-u\|_2^2$. The downstream decision loss instead
measures the utility lost by the induced action:
\begin{equation}
    L_{\mathrm{dec}}(\widehat u,u;s)
    =u_{a^*(s)}(s)-u_{\widehat a(s)}(s).
    \label{eq:static-decision-loss}
\end{equation}

\noindent\textbf{Proposition 1 (Prediction--decision mismatch).}
Prediction loss and decision loss are not equivalent: there exist
predictions $\widehat u^{(1)}$ and $\widehat u^{(2)}$ such that
\begin{equation}
    \begin{aligned}
        L_{\mathrm{pred}}(\widehat u^{(1)},u)
        &<L_{\mathrm{pred}}(\widehat u^{(2)},u),\\
        L_{\mathrm{dec}}(\widehat u^{(1)},u;s)
        &>L_{\mathrm{dec}}(\widehat u^{(2)},u;s).
    \end{aligned}
    \label{eq:prediction-decision-mismatch}
\end{equation}

\noindent\textit{Proof.}
Consider two actions with true utilities $u=(\delta,0)$, where $\delta>0$.
Let $\widehat u^{(1)}=(-\delta,0)$ and
$\widehat u^{(2)}=(\delta+K,K)$ for any $K>\sqrt{2}\delta$. The first
prediction has squared error $4\delta^2$ but reverses the action ordering and
incurs decision loss $\delta$. The second has squared error $2K^2>4\delta^2$ but preserves the optimal action and incurs zero decision loss.
\hfill$\square$

\begin{figure}[!h]
    \centering
    \includegraphics[width=\linewidth]{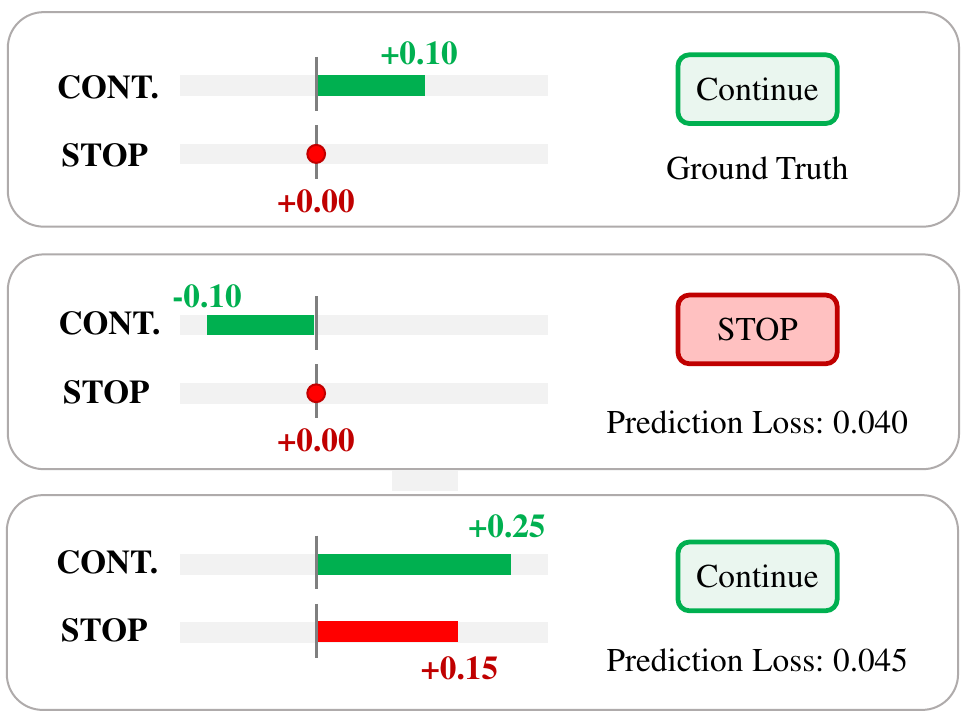}
    \caption{Illustration of the prediction--decision mismatch. A lower prediction loss can reverse the correct decision, whereas a higher prediction loss can preserve it.}
    \label{fig:loss_dismatch}
\end{figure}

Figure~\ref{fig:loss_dismatch} illustrates this construction with $\delta=0.10$ and $K=0.15$: the lower-loss prediction reverses the \textsc{CONTINUE}/\textsc{STOP} ordering, whereas the higher-loss prediction preserves the correct decision. 
Consequently, improving an intermediate quality estimator does not necessarily improve the routing policy it induces. This mismatch motivates learning routing actions from their downstream utility rather than fitting an intermediate quality predictor in isolation. Therefore, we formulate cascade routing as an MDP with a trajectory return that directly captures the objective.

\subsection{MDP Formulation}
\label{sec:mdp}
To jointly model routing and cascading, we formulate model selection and stopping over a candidate LLM pool $\mathcal{M}$ as an MDP:
\begin{equation}
    \mathcal{E} = 
    \left\langle 
    \mathcal{S}, \mathcal{A}, \mathcal{P}, R_{\alpha}, \gamma, T_{\max}
    \right\rangle,
\end{equation}
where $\mathcal{S}$ is the state space, $\mathcal{A}$ is the action space, $\mathcal{P}$ is the transition function, $R_{\alpha}$ is a cost-aware reward function, $\gamma$ is the discount factor, and $T_{\max}$ is the maximum routing depth.

For a query $q$, the state at step $t$ is defined as
\begin{equation}
    s_t = \left(q, y_t, \mathcal{H}_t, C_t, t\right),
\end{equation}
where $y_t$ denotes the current response, $\mathcal{H}_t$ records the previously selected models, $C_t$ is the accumulated inference cost, and $t$ is the current routing depth. At the initial state, no model has been invoked and no response is available.

The set of valid actions is
\begin{equation}
    \mathcal{A}(s_t)
    =
    \{\textsc{Stop}\}
    \cup
    \left(\mathcal{M}\setminus\mathcal{M}_t\right),
\end{equation}
where $\mathcal{M}_t$ is the set of models already selected before step $t$. Because the initial state contains no valid response, the ``stop'' action is masked at $t=0$. Previously selected models are also masked to prevent repeated invocation. Executing a model action $a_t=m_i$ invokes the selected model and updates the current response, model-selection history, accumulated cost, and routing depth. If $a_t=\textsc{Stop}$, the process terminates and returns $y_t$ as the final answer. An episode also terminates when the maximum routing depth $T_{\max}$ is reached.

We train a parameterized policy $\pi_\theta(a_t\mid s_t)$ to maximize the expected cumulative utility over the routing process:
\begin{equation}
    \max_{\theta}\;
    \mathbb{E}_{q\sim\mathcal{D},\,\tau\sim\pi_\theta}
    \left[
        \sum_{t=0}^{T-1}
        \gamma^t R_{\alpha}(s_t,a_t,s_{t+1})
    \right],
    \label{eq:rl-objective}
\end{equation}
where $\tau$ denotes a routing trajectory, $T\leq T_{\max}$ is its adaptive horizon, and $R_{\alpha}$ measures the performance-cost utility of each transition under preference coefficient $\alpha$. 

For a query $q$, let $P_t$ denote the benchmark score of the current response after step $t$, and let $C_t$ denote the cumulative API cost of all model calls up to that step. We normalize performance and cost as
\begin{align}
    \widehat P_t
    &=
    \frac{P_t-P_{\min}(q)}{P_{\max}(q)-P_{\min}(q)},\\
    \widehat C_t
    &=
    1-\frac{C_t}{C_{\max}(q)},
\end{align}
where $P_{\min}(q)$ and $P_{\max}(q)$ are the minimum and maximum scores
obtained by the candidate LLMs on query $q$, respectively. $C_{\max}(q)$ is the cost of invoking the most expensive models allowed by the maximum depth.

The performance-cost utility is defined as
\begin{equation}
    U_{\alpha}(s_t)
    =
    \alpha\,\widehat Q_t
    +(1-\alpha)\,\widehat C_t.
\end{equation}
The policy is trained using the incremental reward
\begin{equation}
    R_{\alpha}
    =
    U_{\alpha}(s_{t+1})-U_{\alpha}(s_t),
\end{equation}
so that, with $\gamma=1$, the cumulative reward of a trajectory equals its final normalized utility.

This formulation enables RLCascadeRouter to jointly learn model selection and stopping, choosing at each step between returning the current response and invoking any previously unselected model. By optimizing performance-cost utility, the policy adapts the model order and cascade depth to each query, learns whether the expected benefit of another selection justifies its cost, and requires neither a quality estimator nor a stopping threshold at inference time.

\subsection{Cascade Policy Network}
\label{sec:hierarchical-policy}
% The Cascade Policy Network serves as a trainable approximator of the stochastic policy $\pi_\theta$ defined by the MDP in Section~\ref{sec:mdp}. Given the state features and valid actions, it produces a raw logit $L_\theta(s_t,a)$ for each action
% \begin{equation}
%     \pi_\theta(a_t\mid s_t)
%     =
%     \frac{
%         \exp\!\left(L_\theta(s_t,a_t)\right)
%     }{
%         \sum_{a'\in\mathcal A}
%         \exp\!\left(L_\theta(s_t,a')\right)
%     }.
%     \label{eq:masked-routing-policy}
% \end{equation}

\begin{figure*}[!ht]
    \centering
    \includegraphics[width=\linewidth]{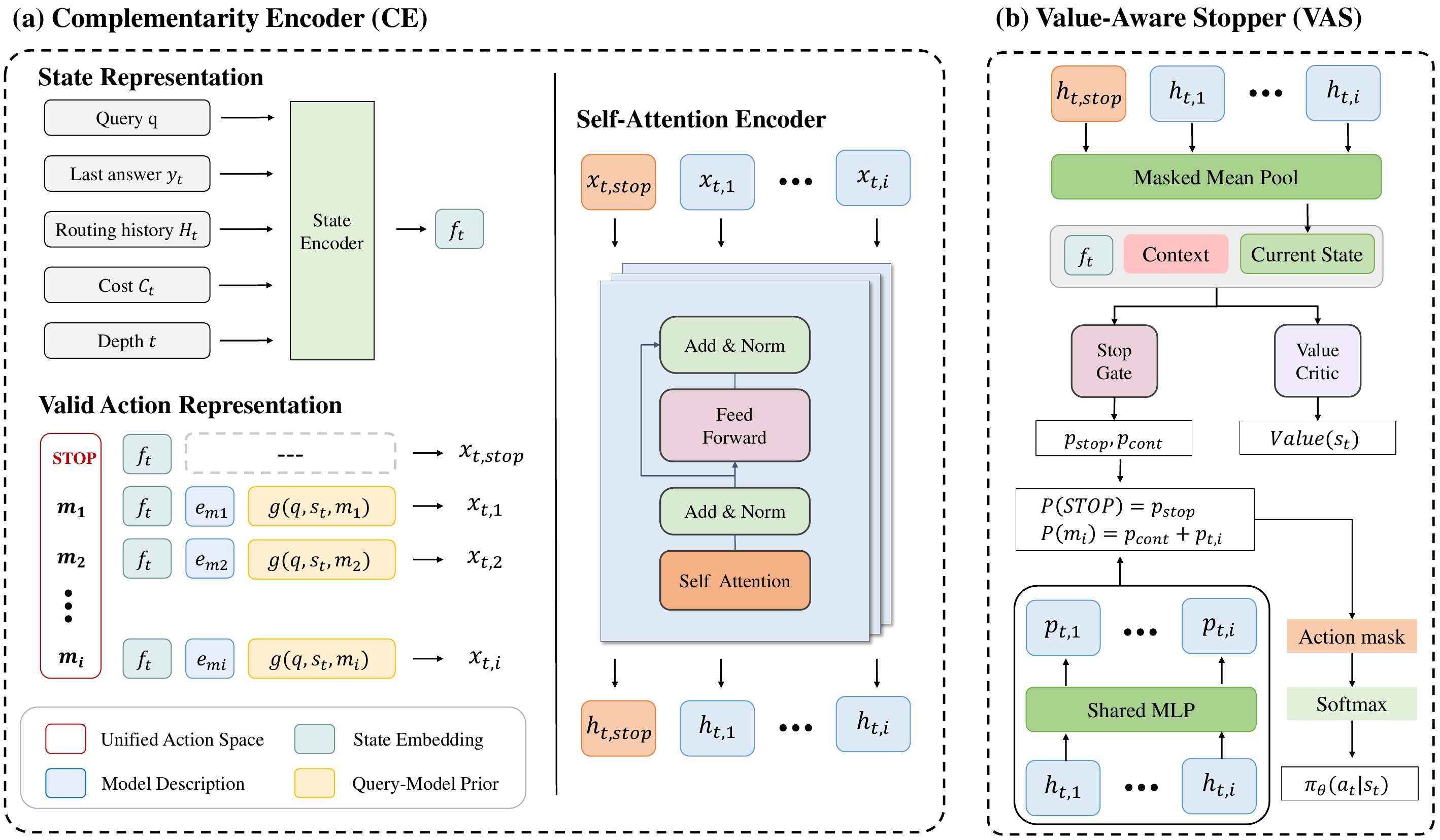}
    \caption{Cascade Policy Network. (a) The Complementarity Encoder (CE) contextualizes state-conditioned action embeddings and scores the remaining models. (b) The Value-Aware Stopper (VAS) aggregates the action context to estimate the STOP/CONTINUE preference and state value. Their outputs form the unified masked routing policy.}
    \label{fig:policy_network}
\end{figure*}

The Bellman structure of the MDP exposes the two comparisons that the
policy must perform. Let $J_\alpha^*(s_t)$ denote the optimal terminal utility
from state $s_t$. Conceptually,
\begin{equation}
    J_\alpha^*(s_t)
    =\max\!\left\{
        U_\alpha(s_t),
        \max_{m_i\in\mathcal M\setminus\mathcal M_t}
        \mathbb E\!\left[J_\alpha^*(s_{t+1})\mid s_t,m_i\right]
    \right\}.
    \label{eq:stop-continue-bellman}
\end{equation}
The inner maximization selects the continuation model with the greatest complementary value, whereas the outer maximization compares the current terminal value with the value attainable by continuing. %To realize these maximizations, 
The Cascade Policy Network employs two jointly trained modules: the Complementarity Encoder (CE) models interactions among the remaining candidates to identify the most valuable continuation, while the Value-Aware Stopper (VAS) evaluates the current state to decide between ``stop'' and model selection. Both modules are optimized using reinforcement-learning feedback, enabling routing without an independent quality estimator. The overall architecture is illustrated in Figure~\ref{fig:policy_network}. We describe its two key components below, with detailed feature definitions and network equations deferred to Appendix~A.

\paragraph{Complementarity Encoder.}The state encoder first maps the query, last answer, routing history, accumulated cost, and cascade depth into a state representation. Each valid model action is then represented by an action token combining the state representation, its model embedding, and query--model features, while ``stop'' is represented by a dedicated action token. The CE jointly processes the valid-action set through stacked self-attention and feed-forward layers. Consequently, each action representation depends on both the routing state and the remaining alternatives, allowing the policy to capture capability complementarity, competition, and redundancy among candidate models.

\paragraph{Value-Aware Stopper.}The VAS applies masked mean pooling to the contextualized valid-action representation and combines the resulting remaining-action context with the state representation. The Stop Gate uses this joint representation to produce \textsc{STOP}/\textsc{CONTINUE} logits, while the value critic estimates the expected future return. The continuation logit is combined with the candidate-model scores, and an action mask followed by softmax yields a unified distribution over ``stop'' and all valid model actions.

\begin{table*}[!ht]
\centering
\caption{Performance comparison of individual LLMs and routing methods in the
performance-cost setting of LLMRouterBench. Bold and gray-shaded values denote
the best and second-best results in each column.}
\label{tab:main-results}
\resizebox{\textwidth}{!}{
\begin{tabular}{lcccccccccccc}
\toprule
& \multicolumn{2}{c}{\textbf{Mathematics}}
& \multicolumn{2}{c}{\textbf{Code}}
& \multicolumn{4}{c}{\textbf{Knowledge}}
& \textbf{IF} & \textbf{Tool}
& \multicolumn{2}{c}{\textbf{Overall}} \\
\cmidrule(lr){2-3}\cmidrule(lr){4-5}\cmidrule(lr){6-9}
\cmidrule(lr){10-10}\cmidrule(lr){11-11}\cmidrule(lr){12-13}
\textbf{Method} & \textbf{AIME} & \textbf{LMB} & \textbf{LCB}
& \textbf{SWE} & \textbf{GPQA} & \textbf{HLE} & \textbf{MMLU-Pro}
& \textbf{SimpleQA} & \textbf{ArenaHard} & \textbf{Tau2}
& \textbf{Avg.}$\uparrow$ & \textbf{Cost}$\downarrow$ \\
\midrule
\multicolumn{13}{l}{\textit{Single-model baselines}} \\
GPT-5 & \cellcolor{gray!20}83.33 & \textbf{78.38}
& \textbf{84.54} & 16.00
& \textbf{88.33} & \textbf{25.97} & 87.22 & 48.00
& 70.35 & 69.05 & 65.12 & 124.81 \\
Gemini-2.5-Pro & 77.78 & 40.54 & 76.97 & \textbf{36.00}
& \cellcolor{gray!20}85.00 & 21.02 & \textbf{87.67} & 54.01
& \cellcolor{gray!20}77.21 & 39.29 & 59.55 & 233.97 \\
Qwen3-235B-Thinking & 72.22 & 48.65 & 75.71 & 20.00
& 80.00 & 8.19 & 80.56 & 49.31
& \textbf{78.10} & 50.00 & 56.27 & 36.84 \\
DeepSeek-R1-0528 & 72.22 & 72.97 & 76.03 & 25.33
& 78.33 & 17.00 & 84.67 & 28.66
& 65.93 & 34.52 & 55.57 & 60.75 \\
GLM-4.6 & \textbf{88.89} & 64.86 & 58.99 & 22.67
& 80.00 & 15.46 & 80.89 & 25.89
& 68.36 & 48.81 & 55.48 & 61.39 \\
Kimi-K2-0905 & 72.22 & \cellcolor{gray!20}75.68 & 62.15 & 24.00
& 71.67 & 4.95 & 80.78 & 30.66
& 74.34 & 51.19 & 54.76 & 14.53 \\
Qwen3-235B & 77.78 & 72.97 & 58.36 & 16.67
& 55.00 & 9.74 & 83.78 & 54.01
& 76.99 & 41.67 & 54.70 & \cellcolor{gray!20}6.98 \\
DeepSeek-V3.1-Terminus & 55.56 & 67.57 & 64.67 & 26.00
& 78.33 & 8.66 & 84.56 & 25.12
& 65.04 & 41.67 & 51.72 & 9.43 \\
Claude-Sonnet-4 & 33.33 & 56.76 & 52.05
& \cellcolor{gray!20}35.33
& 75.00 & 5.10 & 84.11 & 14.02
& 57.52 & 51.19 & 46.44 & 70.28 \\
GPT-5-Chat & 72.22 & 51.35 & 56.78 & 6.67
& 73.33 & 6.03 & 82.33 & 39.29
& 69.47 & $0.00^{*}$ & 45.75 & 17.96 \\
Gemini-2.5-Flash & 55.56 & 64.86 & 55.21 & 20.00
& 58.33 & 6.34 & 81.22 & 30.43
& 56.86 & 28.57 & 45.74 & 32.97 \\
Intern-S1 & 38.89 & 59.46 & 46.69 & 8.00
& 70.00 & 10.36 & 83.00 & 14.33
& 67.48 & 32.14 & 43.03 & 13.85 \\
DeepSeek-V3-0324 & 38.89 & 59.46 & 61.51 & 24.00
& 68.33 & 3.40 & 78.44 & 26.43
& 58.41 & 7.14 & 42.60 & \textbf{4.78} \\
\midrule
\multicolumn{13}{l}{\textit{Routing methods}} \\
Random Router & 61.11 & 64.86 & 64.04 & 18.67
& 73.33 & 10.51 & 82.56 & 35.52
& 68.14 & 39.29 & 51.81 & 50.17 \\
HybridLLM & \cellcolor{gray!20}83.33 & \textbf{78.38}
& \textbf{84.54} & 16.00
& \textbf{88.33} & \textbf{25.97} & 87.22 & 48.00
& 70.35 & 69.05 & 65.12 & 124.97 \\
FrugalGPT & \cellcolor{gray!20}83.33 & \textbf{78.38}
& 81.70 & 26.67
& 81.67 & 22.72 & 84.00 & \cellcolor{gray!20}57.55
& 72.57 & \textbf{75.00} & 66.36 & 124.31 \\
GraphRouter & \cellcolor{gray!20}83.33 & \textbf{78.38}
& \textbf{84.54} & 30.00
& \textbf{88.33} & \cellcolor{gray!20}25.46 & 86.11 & 48.00
& 70.44 & 69.05 & 66.37 & 125.88 \\
Avengers-Pro & \cellcolor{gray!20}83.33 & \textbf{78.38}
& \textbf{84.54} & \textbf{36.00}
& \textbf{88.33} & \textbf{25.97}
& \cellcolor{gray!20}87.44 & 54.78
& 70.58 & 69.05 & \cellcolor{gray!20}67.84 & 135.65 \\
\midrule
\textbf{RLCascadeRouter}
& \cellcolor{gray!20}83.33 & \textbf{78.38}
& \cellcolor{gray!20}84.23 & \textbf{36.00}
& \textbf{88.33} & \cellcolor{gray!20}25.46 & 86.78
& \textbf{58.40} & 76.67 & \cellcolor{gray!20}73.81
& \textbf{68.81} & 180.22 \\
\bottomrule
\end{tabular}}
\end{table*}

\begin{figure*}[!ht]
    \centering
    \includegraphics[width=0.9\linewidth]{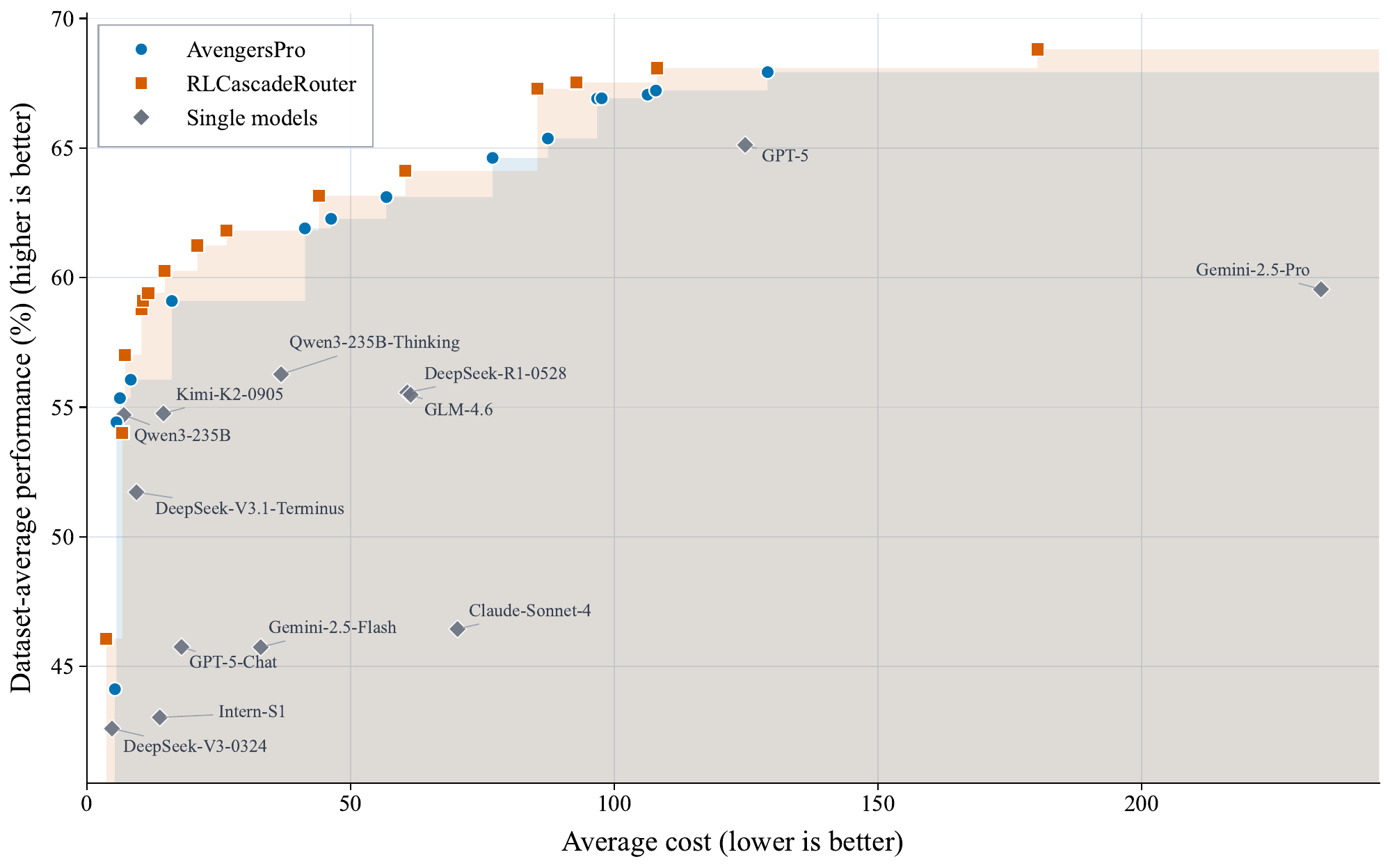}
    \caption {The non-dominated operating points of RLCascadeRouter and Avengers-Pro, together with the individual LLMs.}
    \label{fig:performance_cost}
\end{figure*}

\begin{table*}[!ht]
\centering
\caption{Generalization to unseen models.}
\label{tab:unseen-generalization}
\resizebox{\textwidth}{!}{
\begin{tabular}{lcccccccccccc}
\toprule
Setting & AIME & LMB & LCB & SWE & GPQA & HLE & MMLU-Pro
& SimpleQA & ArenaHard & Avg. & Cost & Unseen Selection \\
\midrule
A: Initial pool
& \textbf{87.50} & \textbf{75.00} & \textbf{93.42} & 23.53
& \textbf{90.57} & \textbf{26.08} & 88.16 & 51.93
& \textbf{77.11} & 59.32 & 82.74 & 0.00 \\

B: Unseen replacement
& 75.00 & 64.29 & 81.58 & 21.01
& 66.04 & 11.42 & 85.51 & 50.46
& 74.67 & 53.97 & \textbf{72.76} & \textbf{98.18} \\

C: Replacement training
& \textbf{87.50} & 71.43 & \textbf{93.42} & \textbf{35.29}
& 86.79 & 19.75 & \textbf{88.65} & \textbf{57.47}
& 76.00 & \textbf{60.61} & 174.76 & 97.59 \\
\bottomrule
\end{tabular}}
\end{table*}

\section{Experimental Setup}
\paragraph{Datasets and Metrics.}
We conduct our experiments on the performance-cost setting of
LLMRouterBench~\cite{li-etal-2026-llmrouterbench}, which contains ten benchmarks:
American Invitational Mathematics Examination (AIME) and LiveMathBench (LMB)
for mathematical reasoning; LiveCodeBench (LCB) and SWE-bench (SWE) for code
generation and software engineering; Graduate-Level Google-Proof Question
Answering (GPQA), Humanity's Last Exam (HLE), MMLU-Pro, and SimpleQA (SQA)
for knowledge and general reasoning; ArenaHard for Instruction Following (IF); and $\tau^2$-Bench (Tau2) for tool use.

We evaluate each routing method from two Metrics: Performance and Cost. To measure Performance, we score the response from the last selected model. We compute the mean score within each dataset and then report the macro-average over the ten datasets. To measure cost, we accumulate the API cost of every model invoked along the complete cascade routing trajectory, including all intermediate calls made before termination.

\paragraph{Baselines.}
For a comprehensive evaluation, we compare RLCascadeRouter with several baselines:

\textbf{Single-model baselines.}
We evaluate each of the 13 candidate LLMs independently on all test queries:
GPT-5, GPT-5-Chat, Gemini-2.5-Pro, Gemini-2.5-Flash,
Claude-Sonnet-4, Qwen3-235B-A22B-2507,
Qwen3-235B-A22B-Thinking-2507, DeepSeek-R1-0528,
DeepSeek-V3-0324, DeepSeek-V3.1-Terminus, GLM-4.6,
Kimi-K2-0905, and Intern-S1. These results characterize the strengths and
costs of individual models and provide the best single-model reference.
%Best Single reference.

\textbf{LLM routing baselines.}
We compare our framework against five representative routing strategies:
(1) \textbf{Random Router}, which uniformly samples one model from the
candidate pool for each query;
(2) \textbf{HybridLLM}~\cite{ding2024hybridllm}, a quality-aware binary
router that predicts query difficulty and routes between the cost-efficient
Qwen3-235B-A22B-2507 and the stronger GPT-5;
(3) \textbf{FrugalGPT}~\cite{chen2024frugalgpt}, which learns an answer
quality scorer and sequentially cascades models according to learned
acceptance thresholds;
(4) \textbf{GraphRouter}~\cite{feng2025graphrouter}, which represents tasks,
queries, and candidate LLMs as a heterogeneous graph and learns their
interactions for model selection; and
(5) \textbf{Avengers-Pro}~\cite{zhang2025avengerspro}, a strong
performance-cost baseline that clusters query embeddings and selects a
model using cluster-conditioned performance and cost statistics.
All embedding-dependent baselines use Qwen3-Embedding-8B~\cite{qwen3embedding} to ensure a consistent representation backbone.

\paragraph{Implementation Details.}
The policy network uses a hidden dimension of 256 and a six-layer Transformer encoder with eight attention heads, a feed-forward dimension of 512, and a dropout rate of 0.1. We train the policy using PPO for 100 iterations with 2048 episodes per iteration. Each iteration performs four optimization epochs with a batch size of 256. We use AdamW with a learning rate of $3\times10^{-5}$ and a weight decay of $10^{-4}$. The PPO clipping coefficient is 0.2, the discount factor is 1.0, the value-loss coefficient is 0.5, and the gradient norm is clipped to 1.0. The entropy coefficient is annealed from 0.03 to 0.005 during training. Within each dataset, queries are randomly divided into 60\% training, 10\% validation, and 30\% testing subsets using random seed 42. RLCascadeRouter allows a maximum of three model calls per query. All experiments are implemented in PyTorch and conducted on an NVIDIA RTX 50-series GPU with 16GB of VRAM.

\section{Experimental Analysis}
\label{sec:experimental-analysis}
% In this section, we present a comprehensive experimental analysis to evaluate the effectiveness of RLCascadeRouter. We first compare its overall performance with representative routing methods and individual LLMs across ten benchmarks (Section~\ref{sec:overall-performance}). We then characterize its performance-cost trade-off under different values of $\alpha$ (Section~\ref{sec:performance-cost-tradeoff}) and evaluate whether our policy can incorporate previously unseen LLMs without retraining (Section~\ref{sec:unseen-model-generalization}). Finally, we conduct controlled ablations (Section~\ref{sec:architecture-ablation}).

\subsection{Overall Performance}
\label{sec:overall-performance}

Table~\ref{tab:main-results} compares RLCascadeRouter with five routing methods and all
13 individual LLMs in the candidate pool. With $\alpha=1.0$, RLCascadeRouter achieves the highest macro-average of $68.81\%$. It improves over Avengers-Pro, the strongest routing baseline, by $0.97$ percentage points, and over GPT-5, the strongest individual model, by $3.69$ points. This result demonstrates that the learned cascade achieves a stronger aggregate result than either one-step routing or fixed single-model deployment. The gains are concentrated on a subset of tasks. Among routing methods, RLCascadeRouter performs best on SimpleQA and ArenaHard, improving over Avengers-Pro by $3.62$ and $6.09$ points, respectively, and it improves Tau2 by $4.76$ points. It also matches the best routing score on AIME, LMB, SWE, and GPQA. While its total cost of $180.22$ is higher than that of Avengers-Pro and GPT-5; the subsequent experiment evaluates the complete Pareto frontier rather than a single high-performance operating point.

\subsection{performance-cost Trade-off} \label{sec:performance-cost-tradeoff}
We vary $\alpha\in\{0,0.05,\ldots,1.0\}$ and train an independent policy at each value. Figure~\ref{fig:performance_cost} shows the non-dominated points of RLCascadeRouter and Avengers-Pro, together with the individual LLMs. Avengers-Pro is the strongest disclosed baseline in LLMRouterBench~\cite{li-etal-2026-llmrouterbench}. RLCascadeRouter provides a more favorable frontier over most of the medium- and high-performance range. Near $59\%$ performance, it reaches $59.11\%$ at a
total cost of $10.64$, compared with $59.10\%$ at $16.12$ for Avengers-Pro, a $34.0\%$ cost reduction. Near $67\%$, it reaches $67.29\%$ at $85.41$, compared
with $67.22\%$ at $107.88$, reducing cost by $20.8\%$. The same RLCascadeRouter operating point also exceeds GPT-5 by $2.17$ points while reducing cost by $31.6\%$. At the upper end of the frontier, RLCascadeRouter reaches $68.08\%$ at
$108.07$, exceeding the maximum performance attained by Avengers-Pro in this
sweep ($67.93\%$). These comparisons confirm that the performance gains shown in Table~\ref{tab:main-results} are not restricted to high-cost scenarios.
%is not limited to the expensive scenario.

\subsection{Generalization} %to Unseen Models}
\label{sec:unseen-model-generalization}
RLCascadeRouter uses textual model descriptions to route unseen LLMs without policy retraining; the descriptions are provided in Appendix~B. Setting~A trains the policy with the original model pool and evaluates it using the same pool. Setting~B evaluates the checkpoint from Setting~A on a modified pool, in which GPT-5, Qwen3-235B-Thinking, and DeepSeek-R1 are replaced by Gemini-2.5-Pro, Claude-Sonnet-4, and Qwen3-235B. Setting~C trains and evaluates the policy on the modified pool. All settings use $\alpha=1.0$.

As shown in Table~\ref{tab:unseen-generalization}, three observations demonstrate the generalization capability of RLCascadeRouter. First, after replacing the three most frequently selected seen models, Setting~B achieves an average performance of $53.97\%$, retaining $90.99\%$ of the $59.32\%$ performance obtained with the initial pool, despite using the original policy without retraining. Second, the replacement models are selected for $98.18\%$ of the queries, indicating that the router can actively incorporate newly introduced candidates rather than simply relying on the remaining seen models. Third, direct training on the replacement pool (Setting~C) yields $60.61\%$ average performance. Without retraining, Setting~B retains $89.05\%$ of this performance and invokes the replacement models at a comparable rate ($98.18\%$ vs.\ $97.59\%$). These results show that RLCascadeRouter can immediately utilize unseen models and preserve most of the achievable routing performance without policy retraining.

\subsection{Ablation Study}
\label{sec:architecture-ablation}

We evaluate the two modules of the Cascade Policy Network.
\textbf{w/o CE} replaces the encoder with independent feed-forward blocks. \textbf{w/o VAS} removes the pooled representation of the remaining actions from the stopping branch, so that STOP/CONTINUE depends only on the current state. In addition to these component ablations, we include \textbf{QE Stop} as a stopping control. It replaces the policy-based STOP/CONTINUE decision with a post-hoc quality estimator and a calibrated threshold while retaining the Full model selector. We report three representative preferences: $\alpha=1.0$, $0.5$, and $0.2$.

\begin{table}[h]
\centering
\caption{Policy ablations and evaluator-based stopping control under three performance-cost preferences. Bold denotes the best result in each column.}
\label{tab:architecture-ablation}
\resizebox{\columnwidth}{!}{
\begin{tabular}{lrrrrrr}
\toprule
\multirow{2}{*}{Variant}
& \multicolumn{2}{c}{$\alpha=1.0$}
& \multicolumn{2}{c}{$\alpha=0.5$}
& \multicolumn{2}{c}{$\alpha=0.2$} \\
\cmidrule(lr){2-3}\cmidrule(lr){4-5}\cmidrule(lr){6-7}
& Perf.$\uparrow$ & Cost$\downarrow$
& Perf.$\uparrow$ & Cost$\downarrow$
& Perf.$\uparrow$ & Cost$\downarrow$ \\
\midrule
Full
& \textbf{68.81} & 180.22
& \textbf{61.91} & 28.06
& 58.02 & 10.13 \\
w/o CE
& 68.30 & 152.60
& 60.32 & 21.05
& 56.81 & \textbf{7.23} \\
w/o VAS
& 67.43 & 191.56
& 60.67 & \textbf{15.80}
& \textbf{58.62} & 9.97 \\
\midrule
QE Stop
& 67.70 & \textbf{135.76}
& 61.23 & 27.43
& 58.02 & 10.13 \\
\bottomrule
\end{tabular}}
\end{table}

\textbf{w/o CE.} Full outperforms w/o CE by $0.51$, $1.59$, and $1.21$ points at $\alpha=1.0$, $0.5$, and $0.2$, respectively, supporting the need to model complementarity among candidate models. \textbf{w/o VAS.} Removing VAS reduces performance by $1.38$ and $1.23$ points at $\alpha=1.0$ and $0.5$, but improves it by $0.60$ points at $\alpha=0.2$, indicating that remaining-action context is primarily beneficial in multi-step regimes. \textbf{QE Stop.} Replacing learned stopping with a quality-estimator threshold reduces performance by $1.11$ and $0.67$ points at $\alpha=1.0$ and $0.5$. The methods are identical at $\alpha=0.2$, where both reduce to single-step routing, showing that utility-based stopping is most effective when continuation is actively considered.

\section{Conclusion}
In this work, we present RLCascadeRouter, a quality-estimator-free cascade router that replaces predict-then-optimize quality estimation with a decision-aligned MDP policy. Its Cascade Policy Network jointly learns candidate complementarity for model selection and remaining-action value for stopping. Experiments on ten benchmarks demonstrate strong performance-cost trade-offs, generalization to unseen models without retraining, and the effectiveness of the proposed policy design. Future work may extend this framework to online environments with dynamically changing model pools, costs, latencies, and response distributions.

% \clearpage
\bibliographystyle{named}
\bibliography{ref}

\clearpage
\onecolumn
\appendix
\section{Cascade Policy Network Details}
\label{app:policy-network-details}

This section provides the complete feature definitions and network computations of the Cascade Policy Network (CPN). The CPN parameterizes the stochastic policy $\pi_\theta(a_t\mid s_t)$ and the value function $V_\phi(s_t)$ of the routing MDP. It consists of a Complementarity Encoder (CE), which compares the remaining model actions, and a Value-Aware Stopper (VAS), which determines whether the best continuation opportunity justifies an additional model call.

\subsection{State Representation}
A query-only state representation cannot distinguish an unanswered query from one that already has a strong response, nor can it describe previous model calls and their accumulated costs. Let $E(\cdot)$ denote the embedding encoder. After $t$ model calls, we construct the state representation as follows:
\begin{equation}
    f_t=
    [e_q;e_{y_t};\bar e_{\mathcal M_t};
     e_{m_t};z_t;d_q],
    \label{eq:app-state-representation}
\end{equation}
where $e_q=E(q)$ and $e_{y_t}=E(y_t)$ encode the query and the current response, respectively. The aggregated representation of selected models is defined as
% The selected-model representation is
\begin{equation}
    \bar e_{\mathcal M_t}
    =
    \frac{1}{|\mathcal M_t|}
    \sum_{m_j\in\mathcal M_t}e_{m_j},
\end{equation}
where $e_{m_j}$ is obtained by embedding the textual description of candidate %descriptor of 
model $m_j$. The feature $e_{m_t}$ represents the most recently invoked model, and $d_q$ is the dataset indicator. The scalar routing features are
\begin{equation}
    z_t=
    \left[
        \log(1+C_t);
        \frac{t}{T_{\max}};
        \mathbb I(t>0);
        \frac{|\mathcal M_t|}{|\mathcal M|}
    \right],
    \label{eq:app-scalar-state}
\end{equation}
which encode the accumulated cost, the normalized cascade depth, response availability, and the fraction of selected models. In the initial state, $e_{y_t}$, $\bar e_{\mathcal M_t}$, and $e_{m_t}$ are set to zero vectors.

\subsection{Query-Model Prior}
Model descriptions provide semantic information but do not directly describe how individual models perform on different query types. We therefore construct a query-conditioned prior exclusively from the training split. Given a query embedding $e_q$ and $K$ query-type centroids $\{\mu_k\}_{k=1}^{K}$, the soft membership of query $q$ in type $k$ is defined as
\begin{equation}
    w_k(q)
    =
    \frac{
        \exp\!\left(
            \operatorname{cos}(e_q,\mu_k)/\tau
        \right)
    }{
        \sum_{k'=1}^{K}
        \exp\!\left(
            \operatorname{cos}(e_q,\mu_{k'})/\tau
        \right)
    },
    \label{eq:app-query-membership}
\end{equation}
where $\tau$ is a temperature parameter. Let $\varphi_{k,i}$ denote the smoothed performance-cost statistics of model $m_i$ for query type $k$, including its performance, accuracy, cost, score percentile, and observation count. The Query-Model Prior is then computed as
\begin{equation}
    p(q,m_i)
    =
    \sum_{k=1}^{K}w_k(q)\varphi_{k,i}.
    \label{eq:app-query-model-prior}
\end{equation}
This prior is supplied as an input feature rather than used as an external routing rule.

\subsection{Action Representation}
Let
\begin{equation}
    \mathcal A
    =
    \{\textsc{STOP}\}\cup\mathcal M
\end{equation}
denote the complete action universe. The validity mask $\mu_t(a)\in\{0,1\}$ is defined as
\begin{equation}
    \mu_t(a)=
    \begin{cases}
        \mathbb I(t>0),
        & a=\textsc{STOP},\\
        \mathbb I(m_i\notin\mathcal M_t)
        \mathbb I(t<T_{\max}),
        & a=m_i.
    \end{cases}
    \label{eq:app-action-mask}
\end{equation}
Thus, \textsc{STOP} is unavailable before the first response, while previously selected models and further model calls at the maximum depth are masked.

For each valid model action $m_i$, we construct the auxiliary feature vector as follows:
\begin{equation}
\begin{aligned}
    g(q,s_t,m_i)
    =[
        &\widetilde c(q,m_i);
        \operatorname{cos}(e_q,e_{m_i});\\
        &\operatorname{cos}(e_{y_t},e_{m_i});
        \operatorname{cos}(\bar e_{\mathcal M_t},e_{m_i});\\
        &p(q,m_i)
    ],
\end{aligned}
    \label{eq:app-query-model-features}
\end{equation}
where $\widetilde c(q,m_i)$ is the normalized expected invocation cost. Similarities involving unavailable response or history features are set to zero. The corresponding action token is defined as
\begin{equation}
    x_{t,i}
    =
    [f_t;e_{m_i};g(q,s_t,m_i);0],
    \label{eq:app-model-action-token}
\end{equation}
where the final scalar is the action-type indicator. The \textsc{STOP} token uses the same state representation but contains no model-specific features:
\begin{equation}
    x_{t,\mathrm{stop}}
    =
    [f_t;\mathbf 0;\mathbf 0;1].
    \label{eq:app-stop-action-token}
\end{equation}
This token-based construction allows all candidate models to be scored using shared parameters rather than model-index-specific output heads.

\subsection{Complementarity Encoder}
The marginal value of a model depends on the current response, the models already selected, and the alternatives that remain available. The CE therefore contextualizes every action against the complete valid-action set. Each action token is first projected into a shared hidden space and combined with a state projection as follows:
\begin{equation}
    h_{t,a}^{(0)}
    =
    P_a(x_{t,a})+P_s(f_t),
    \qquad a\in\mathcal A_t,
    \label{eq:app-ce-input}
\end{equation}
where $\mathcal A_t=\{a\in\mathcal A:\mu_t(a)=1\}$. Let
$H_t^{(\ell)}$ denote the matrix containing the valid-action representations at layer $\ell$. Each of the $L$ Transformer blocks computes the following:
\begin{align}
    \widetilde H_t^{(\ell)}
    &=
    \operatorname{LN}\!\left(
        H_t^{(\ell-1)}
        +
        \operatorname{MHA}
        \!\left(H_t^{(\ell-1)}\right)
    \right),
    \label{eq:app-ce-attention}\\
    H_t^{(\ell)}
    &=
    \operatorname{LN}\!\left(
        \widetilde H_t^{(\ell)}
        +
        \operatorname{FFN}
        \!\left(\widetilde H_t^{(\ell)}\right)
    \right),
    \label{eq:app-ce-ffn}
\end{align}
where $\operatorname{MHA}$ denotes masked multi-head self-attention and
\begin{equation}
    \operatorname{FFN}(h)
    =
    W_2\operatorname{GELU}(W_1h+b_1)+b_2.
\end{equation}
The final contextualized representation of action $a$ is
$h_{t,a}=H_{t,a}^{(L)}$. Because self-attention operates over the complete valid-action set, $h_{t,i}$ represents the relative contribution of model $m_i$ rather than an isolated compatibility score. A shared model actor produces its continuation score as follows:
\begin{equation}
    \rho_{t,i}
    =
    \operatorname{ModelActor}(h_{t,i}),
    \qquad m_i\in\mathcal A_t.
    \label{eq:app-model-score}
\end{equation}
Sharing the actor across model tokens also avoids tying the policy output to fixed model indices.

\subsection{Value-Aware Stopper}
Stopping requires comparing the utility already obtained with the opportunities remaining in the action set. The VAS summarizes these opportunities by masked mean pooling as follows:
\begin{equation}
    \bar h_t
    =
    \frac{
        \sum_{a\in\mathcal A}
        \mu_t(a)h_{t,a}
    }{
        \sum_{a\in\mathcal A}\mu_t(a)
    }.
    \label{eq:app-masked-pooling}
\end{equation}
The pooled candidate-set representation is concatenated with the current state:
\begin{equation}
    c_t^{\mathrm{ctx}}
    =
    [f_t;\bar h_t].
    \label{eq:app-vas-context}
\end{equation}
The %stopping gate 
Stop Gate and value critic are then computed as follows:
\begin{align}
    [\ell_t^{\mathrm{stop}},
     \ell_t^{\mathrm{cont}}]
    &=
    \operatorname{StopGate}
    (c_t^{\mathrm{ctx}}),
    \label{eq:app-stop-gate}\\
    V_\phi(s_t)
    &=
    \operatorname{Critic}
    (c_t^{\mathrm{ctx}}).
    \label{eq:app-value-critic}
\end{align}
The critic is used to estimate policy advantages during training and does not directly determine the selected action. Unlike an external response-quality estimator, the VAS does not predict an absolute quality label or apply a calibrated threshold. Its stopping preference is learned jointly with model selection from the trajectory-level performance-cost return.

\subsection{Joint Routing Policy}
The CE and the VAS form a single hierarchical policy. Their outputs are combined into unified action logits as follows:
\begin{align}
    L_t(\textsc{STOP})
    &=
    \ell_t^{\mathrm{stop}},
    \label{eq:app-stop-logit}\\
    L_t(m_i)
    &=
    \ell_t^{\mathrm{cont}}+\rho_{t,i}.
    \label{eq:app-model-logit}
\end{align}
The VAS therefore controls the global preference between stopping and continuing, while the CE determines which model is preferred conditional on continuation. Applying the validity mask %gives
yields the following masked routing policy:
\begin{equation}
    \pi_\theta(a_t\mid s_t,\mu_t)
    =
    \frac{
        \mu_t(a_t)
        \exp\!\left(L_t(a_t)\right)
    }{
        \sum_{a'\in\mathcal A}
        \mu_t(a')
        \exp\!\left(L_t(a')\right)
    }.
    \label{eq:app-masked-routing-policy}
\end{equation}
At inference time, the policy selects an action from this masked distribution and either returns the current response or invokes the selected model. It requires neither ground-truth response scores nor an independent post-hoc quality estimator.

% Requires \usepackage{booktabs}.
% The description text below is reproduced verbatim from
% config/model_descriptions.json.

% \clearpage
\section{Model Descriptions}
\label{app:model-descriptions}

RLCascadeRouter represents each candidate model using an embedding of its
textual description. The following tables provide the exact textual
descriptions of all 13 candidate models used in our experiments.

\newcommand{\modeldescriptiontable}[2]{%
\begin{table*}[!htbp]
\centering
\caption{Description of \textbf{#1}.}
\begin{tabular}{@{}p{\textwidth}@{}}
\toprule
#2 \\
\bottomrule
\end{tabular}
\end{table*}
}

% \subsection{Descriptions of the 13 Candidate Models}

\modeldescriptiontable{Gemini-2.5-Flash}{
Gemini 2.5 Flash is Google's efficiency-oriented multimodal thinking model for low-latency and high-volume workloads. It handles text, images, audio, video, and long-context inputs, and supports configurable thinking, function calling, code execution, and grounded retrieval. The model is suitable for general question answering, document understanding, routine reasoning, and agentic tasks where throughput and cost efficiency are important.
}

\modeldescriptiontable{GPT-5-Chat}{
GPT-5 Chat is OpenAI's non-reasoning GPT-5 model previously used for conversational interaction in ChatGPT. It is designed for direct, responsive instruction following, general knowledge assistance, writing, and multi-turn dialogue, with support for function calling and structured outputs. The model is suitable for conversational and general-purpose tasks that benefit from strong language understanding without the additional computation of extended reasoning.
}

\modeldescriptiontable{GPT-5}{
GPT-5 Medium is OpenAI's GPT-5 reasoning model configured with medium reasoning effort. It is designed for complex reasoning, software coding, mathematics, instruction following, and agentic tool use across domains, and supports long-context inputs, function calling, and structured outputs. The medium setting balances solution quality and inference effort, making it suitable for challenging tasks that require deliberate reasoning without the maximum reasoning budget.
}

\modeldescriptiontable{Qwen3-235B-Thinking}{
Qwen3-235B-A22B-Thinking-2507 is Qwen's mixture-of-experts reasoning model with 235 billion total parameters and 22 billion activated parameters, operating exclusively in thinking mode. It uses an extended reasoning process for logical reasoning, mathematics, science, coding, academic problems, instruction following, and tool use. The model is suitable for highly complex tasks where deeper reasoning and long-context understanding justify greater generation length and inference cost.
}

\modeldescriptiontable{DeepSeek-V3}{
DeepSeek-V3-0324 is DeepSeek's general-purpose mixture-of-experts instruction model operating in a direct, non-thinking mode. It is designed for general question answering, reasoning, software coding, front-end development, instruction following, and tool use. The model is suitable for broad language and coding workloads, especially tasks that require capable and efficient direct generation rather than the longer explicit reasoning process of a dedicated reasoning model.
}

\modeldescriptiontable{DeepSeek-V3.1-Terminus}{
DeepSeek-V3.1-Terminus is DeepSeek's hybrid inference model supporting both thinking and non-thinking behavior. It combines general language and reasoning capabilities with tool use, multi-step search, software coding, and agentic task execution, while improving language consistency and output reliability. The model is suitable for general assistants, code and search agents, and workflows that require flexible reasoning depth with a long context window.
}

\modeldescriptiontable{DeepSeek-R1}{
DeepSeek-R1-0528 is DeepSeek's dedicated reasoning model that generates an explicit reasoning process prior to producing its final answer. It is designed for difficult mathematics, logic, science, software coding, and other multi-step analytical tasks, with support for function calling and structured JSON output. The model is suitable when reasoning depth and solution quality are more important than latency, output length, or inference cost.
}

\modeldescriptiontable{GLM-4.6}{
GLM-4.6 is Z.ai's general-purpose reasoning and agentic model for coding, long-context processing, search, writing, and tool-based workflows. It supports tool use during inference and a 200K-token context window, with a particular emphasis on software engineering, front-end generation, search agents, and readable long-form writing. The model is suitable for complex multi-step tasks that combine reasoning, code, external tools, and substantial context.
}

\modeldescriptiontable{Kimi-K2}{
Kimi-K2-Instruct-0905 is Moonshot AI's mixture-of-experts instruction model with one trillion total parameters and 32 billion activated parameters, operating as a direct-response model without extended thinking. It is optimized for agentic coding, tool calling, front-end development, general chat, and autonomous problem-solving, with a 256K-token context window. The model is suitable for long-horizon coding and agent workflows that require capable, responsive tool use.
}

\modeldescriptiontable{Intern-S1}{
Intern-S1 is InternLM's multimodal mixture-of-experts reasoning model specialized for scientific problem-solving while retaining strong general language and vision capabilities. It is designed for scientific reasoning across chemistry, materials science, life science, and earth science, including molecular formulas, chemical structures, protein sequences, synthesis planning, and physical signals. The model is suitable for research-oriented questions and multimodal scientific tasks requiring specialized domain knowledge.
}

\modeldescriptiontable{Gemini-2.5-Pro}{
Gemini 2.5 Pro is Google's high-capability multimodal thinking model for complex problems in mathematics, science, coding, and analytical reasoning. It handles text, images, audio, video, documents, and extended contexts, and supports function calling, code execution, and grounded retrieval. The model is suitable for difficult multi-step tasks, large document or codebase analysis, and workloads that prioritize answer quality over latency.
}

\modeldescriptiontable{Claude-Sonnet-4}{
Claude Sonnet 4 is Anthropic's hybrid reasoning model with both fast-response and extended-thinking modes. It is designed for precise instruction following, complex reasoning, software coding, agentic workflows, and tool use. The model balances capability and efficiency, making it suitable for general assistants, multi-step problem-solving, codebase navigation, and high-volume tasks that may benefit from deeper reasoning.
}

\modeldescriptiontable{Qwen3-235B}{
Qwen3-235B-A22B-Instruct-2507 is Qwen's mixture-of-experts instruction model with 235 billion total parameters and 22 billion activated parameters, operating exclusively in non-thinking mode. It provides multilingual instruction following, text comprehension, logical reasoning, mathematics, science, coding, tool use, and long-context understanding. The model is suitable for broad general-purpose and open-ended tasks that require strong direct responses without an extended explicit reasoning trace.
}

\let\modeldescriptiontable\relax

\end{document}